\documentclass[10pt,twocolumn,letterpaper]{article}

\usepackage{cvpr}
\usepackage{times}
\usepackage{epsfig}
\usepackage{graphicx}
\usepackage{amsmath}
\usepackage{amssymb}

\usepackage[utf8]{inputenc} 
\usepackage[T1]{fontenc}    
\usepackage{booktabs}       
\usepackage{amsfonts}       
\usepackage{nicefrac}       
\usepackage{microtype}      

\usepackage{amsmath,amssymb} 
\usepackage{color}
\usepackage{algorithm}
\usepackage{algpseudocode}
\usepackage{mathtools} 
\usepackage{booktabs} 
\usepackage{multirow}
\usepackage{subcaption}
\usepackage{hyphenat} 

\usepackage{color}    
\usepackage{listings} 
\usepackage{setspace} 
\usepackage{xparse}   
\usepackage{pdfpages} 

\usepackage[pagebackref=true,breaklinks=true,letterpaper=true,colorlinks,bookmarks=false]{hyperref}

\DeclareMathOperator*{\argmax}{arg\,max}

\DeclarePairedDelimiterX{\infdivx}[2]{(}{)}{%
  #1\delimsize\|#2%
}

\NewDocumentCommand{\codeword}{v}{%
\texttt{#1}%
}

\graphicspath{{./figures/}}

\newcommand{\ignore}[1]{}
\newcommand{\mbf}[1]{\mathbf{#1}}

\DeclareRobustCommand\code[1]{%
  \ifmmode
    \expandafter\texttt
  \else
    \expandafter\textnhtt
  \fi{#1}%
}

\definecolor{Code}{rgb}{0,0,0}
\definecolor{Decorators}{rgb}{0.5,0.5,0.5}
\definecolor{Numbers}{rgb}{0.5,0,0}
\definecolor{MatchingBrackets}{rgb}{0.25,0.5,0.5}
\definecolor{Keywords}{rgb}{0,0,1}
\definecolor{self}{rgb}{0,0,0}
\definecolor{Strings}{rgb}{0,0.63,0}
\definecolor{Comments}{rgb}{0,0.63,1}
\definecolor{Backquotes}{rgb}{0,0,0}
\definecolor{Classname}{rgb}{0,0,0}
\definecolor{FunctionName}{rgb}{0,0,0}
\definecolor{Operators}{rgb}{0,0,0}
\definecolor{Background}{rgb}{0.98,0.98,0.98}

\lstdefinestyle{default}{
    backgroundcolor=\color{Background},   
    commentstyle=\color{Comments},
    keywordstyle=\color{Keywords},
    numberstyle=\tiny\color{Numbers},
    stringstyle=\color{Strings},
    basicstyle=\footnotesize,
    breakatwhitespace=false,       
    breaklines=true,         
    captionpos=b,                 
    keepspaces=true,                 
    numbers=left,
    numberstyle=\footnotesize,
    numbersep=1em,
    xleftmargin=1em,
    framextopmargin=2em,
    framexbottommargin=1em,
    showspaces=false,
    showstringspaces=false,
    showtabs=false,                  
    tabsize=2,
}

\lstdefinelanguage{Python}{
    frame=l,
    basicstyle=\ttfamily\small\setstretch{1},
    commentstyle=\color{Comments}\slshape,
    stringstyle=\color{Strings},
    morecomment=[s][\color{Strings}]{"""}{"""},
    morecomment=[s][\color{Strings}]{'''}{'''},
    morekeywords={import,from,class,def,for,while,if,is,in,elif,else,not,and,or,print,break,continue,return,True,False,None,access,as,del,except,exec,finally,global,import,lambda,pass,print,raise,try,assert},
    keywordstyle={\color{Keywords}\bfseries},
    morekeywords={[2]@invariant,pylab,numpy,np,scipy},
    keywordstyle={[2]\color{Decorators}\slshape},
    emph={self},
    emphstyle={\color{self}\slshape},
}

\lstdefinelanguage{cpp}{
    basicstyle=\footnotesize \ttfamily \color{black} \bfseries,
    breakatwhitespace=false,
    commentstyle=\color{dkgreen},
    deletekeywords={...},
    escapeinside={\%*}{*)},
    language=C++,
    morekeywords={string,cerr,exit},
    identifierstyle=\color{black},
    stringstyle=\color{blue},
    numberstyle=\tiny\color{black},
    rulecolor=\color{black},
    stepnumber=1,
    tabsize=2,
    title=\lstname,         
}

\cvprfinalcopy 

\def\cvprPaperID{1523} 

\ifcvprfinal\fi
\begin{document}

\title{4D Spatio-Temporal ConvNets: Minkowski Convolutional Neural Networks}

%

\author{
  Christopher Choy \\
  \small{\texttt{chrischoy@stanford.edu}}
   \and
   JunYoung Gwak \\
   \small{\texttt{jgwak@stanford.edu}}
   \and
   Silvio Savarese \\
   \small{\texttt{ssilvio@stanford.edu}}
}

\maketitle

\begin{abstract}
In many robotics and VR/AR applications, 3D-videos are readily-available sources of input (a continuous sequence of depth images, or LIDAR scans). However, these 3D-videos are processed frame-by-frame either through 2D convnets or 3D perception algorithms in many cases. In this work, we propose 4-dimensional convolutional neural networks for spatio-temporal perception that can directly process such 3D-videos using high-dimensional convolutions. For this, we adopt sparse tensors~\cite{sparsecnn, sparse3dcnn} and propose the generalized sparse convolution which encompasses all discrete convolutions. To implement the generalized sparse convolution, we create an open-source auto-differentiation library for sparse tensors that provides extensive functions for high-dimensional convolutional neural networks.\footnote{\url{https://github.com/StanfordVL/MinkowskiEngine}} We create 4D spatio-temporal convolutional neural networks using the library and validate them on various 3D semantic segmentation benchmarks and proposed 4D datasets for 3D-video perception. To overcome challenges in the high-dimensional 4D space, we propose the hybrid kernel, a special case of the generalized sparse convolution, and the trilateral-stationary conditional random field that enforces spatio-temporal consistency in the 7D space-time-chroma space. Experimentally, we show that convolutional neural networks with only generalized sparse convolutions can outperform 2D or 2D-3D hybrid methods by a large margin.\footnote{At the time of submission, we achieved the best performance on ScanNet~\cite{scannet} with 67.9\% mIoU} Also, we show that on 3D-videos, 4D spatio-temporal convolutional neural networks are robust to noise, outperform 3D convolutional neural networks and are faster than the 3D counterpart in some cases.

\end{abstract}

\section{Introduction}

\begin{figure}[t]
    \centering
    \begin{tabular}{ccc}
    \includegraphics[width=0.32\columnwidth]{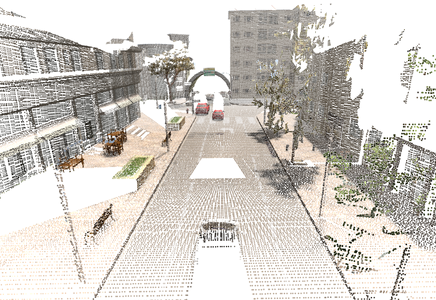}
    \includegraphics[width=0.32\columnwidth]{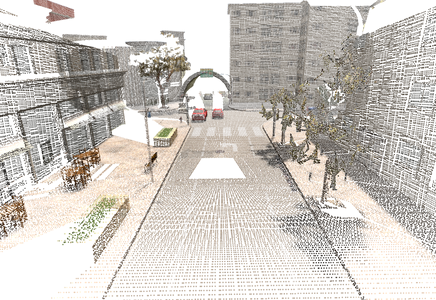}
    \includegraphics[width=0.32\columnwidth]{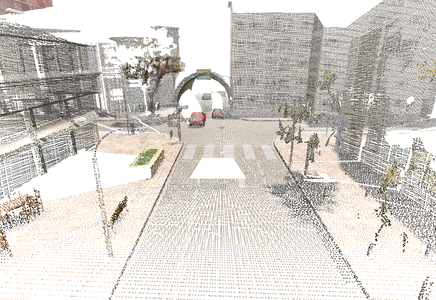}
    \end{tabular}
    \vspace{-0.5em}
    \caption{An example of 3D video: 3D scenes at different time steps. Best viewed on display.}
    \label{fig:3dvideo}
    \vspace{-0.5em}
\end{figure}

In this work, we are interested in 3D-video perception. A 3D-video is a temporal sequence of 3D scans such as a video from a depth camera, a sequence of LIDAR scans, or a multiple MRI scans of the same object or a body part (Fig.~\ref{fig:3dvideo}). As LIDAR scanners and depth cameras become more affordable and widely used for robotics applications, 3D-videos became readily-available sources of input for robotics systems or AR/VR applications.

However, there are many technical challenges in using 3D-videos for high-level perception tasks. First, 3D data requires heterogeneous representations and processing those either alienates users or makes it difficult to integrate into larger systems. Second, the performance of the 3D convolutional neural networks is worse or on-par with 2D convolutional neural networks. Third, there are limited number of open-source libraries for fast large-scale 3D data.

To resolve most, if not all, of the challenges in the high-dimensional perception, we adopt a sparse tensor~\cite{sparsecnn, sparse3dcnn} for our problem and propose the generalized sparse convolutions. The generalized sparse convolution encompasses all discrete convolutions as its subclasses and is crucial for high-dimensional perception. We implement the generalized sparse convolution and all standard neural network functions in Sec.~\ref{sec:minkowskiengine} and open-source the library.

\begin{figure}[t]
    \centering
    \minipage{0.2\columnwidth}
      \includegraphics[width=\linewidth]{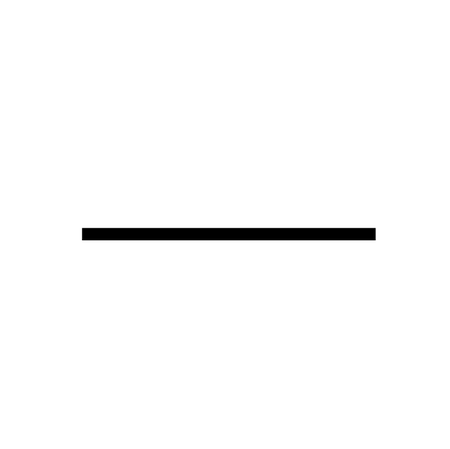}
      \vspace{-2em}
      \caption*{\small{1D: Line}}\label{fig:line}
    \endminipage\hfill
    \minipage{0.2\columnwidth}
      \includegraphics[width=\linewidth]{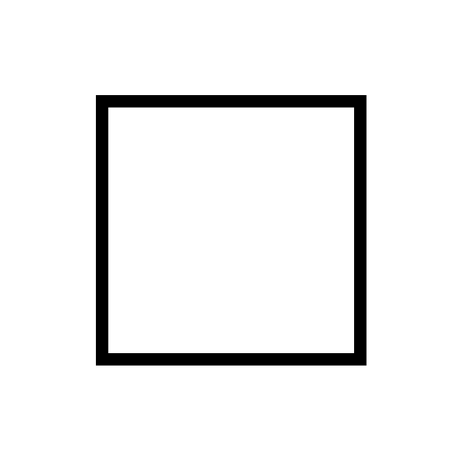}
      \vspace{-2em}
      \caption*{\small{2D: Square}}\label{fig:square}
    \endminipage\hfill
    \minipage{0.2\columnwidth}
      \includegraphics[width=\linewidth]{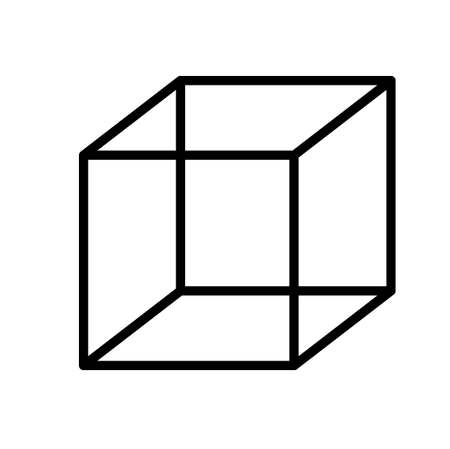}
      \vspace{-2em}
      \caption*{\small{3D: Cube}}\label{fig:cube}
    \endminipage\hfill
    \minipage{0.25\columnwidth}
        \centering
        \includegraphics[width=0.8\linewidth]{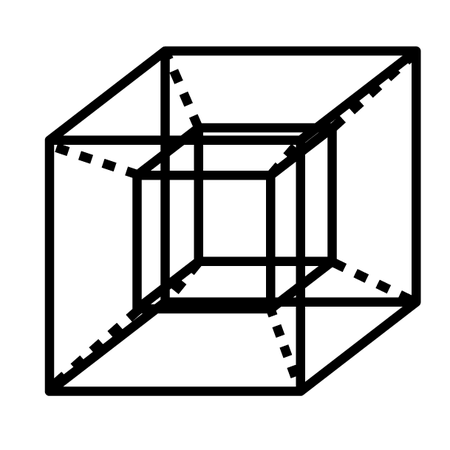}
      \vspace{-0.8em}
      \caption*{\small{4D: Tesseract}}\label{fig:tesseract}
    \endminipage
    \vspace{-0.5em}
    \caption{2D projections of hypercubes in various dimensions}
    \label{fig:hypercubes}
    \vspace{-1.0em}
\end{figure}

We adopt the sparse representation for several reasons. Currently, there are various concurrent works for 3D perception: a dense 3D convolution~\cite{scannet}, pointnet-variants~\cite{pointnet, pointnetpp}, continuous convolutions~\cite{montecarlo, pointcnn}, surface convolutions~\cite{surface, tangent}, and an octree convolution~\cite{octnet}. Out of these representations, we chose a sparse tensor due to its expressiveness and generalizability for high-dimensional spaces. Also, it allows homogeneous data representation within traditional neural network libraries since most of them support sparse tensors.

Second, the sparse convolution closely resembles the standard convolution (Sec.~\ref{sec:sparse}) which is proven to be successful in 2D perception as well as 3D reconstruction~\cite{3dr2n2}, feature learning~\cite{3dmatch}, and semantic segmentation~\cite{segcloud}.

Third, the sparse convolution is efficient and fast. It only computes outputs for predefined coordinates and saves them into a compact sparse tensor (Sec.~\ref{sec:sparse}). It saves both memory and computation especially for 3D scans or high-dimensional data where most of the space is empty.

Thus, we adopt the sparse representation for the our problem and create the first large-scale 3D/4D networks or Minkowski networks.\footnote{At the time of submission, our proposed method was the first very deep 3D convolutional neural networks with more than 20 layers.} We named them after the space-time continuum, Minkowski space, in Physics.

However, even with the efficient representation, merely scaling the 3D convolution to high-dimensional spaces results in significant computational overhead and memory consumption due to the curse of dimensionality. A 2D convolution with kernel size 5 requires $5^2=25$ weights which increases exponentially to $5^3 = 125$ in a 3D cube, and 625 in a 4D tesseract (Fig.~\ref{fig:hypercubes}). This exponential increase, however, does not necessarily lead to better performance and slows down the network significantly. To overcome this challenge, we propose custom kernels with non-(hyper)-cubic shapes using the generalized sparse convolution.

Finally, the predictions from the 4D spatio-temporal generalized sparse convnets are not necessarily consistent throughout the space and time. To enforce consistency, we propose high-dimensional conditional random fields defined in a 7D trilateral space (space-time-color) with a stationary pairwise consistency function. We use variational inference to convert the conditional random field to differentiable recurrent layers which can be implemented in as a 7D generalized sparse convnet and train both the 4D and 7D networks end-to-end.

Experimentally, we use various 3D benchmarks that cover both indoor~\cite{scannet, stanford3dis} and outdoor spaces~\cite{synthia, ruemonge2014}. First, we show that networks with only generalized sparse 3D conv nets can outperform 2D or hybrid deep-learning algorithms by a large margin.\footnote{We achieved 67.9\% mIoU on the ScanNet benchmark outperforming all algorithms including the best peer-reviewed work~\cite{3dmv} by 19\% mIoU at the time of submission.} Also, we create 4D datasets from Synthia~\cite{synthia} and Varcity~\cite{ruemonge2014} and report ablation studies of temporal components. Experimentally, we show that the generalized sparse conv nets with the hybrid kernel outperform sparse convnets with tesseract kernels. Also, the 4D generalized sparse convnets are more robust to noise and sometimes more efficient in some cases than the 3D counterpart.

\section{Related Work}

The 4D spatio-temporal perception fundamentally requires 3D perception as a slice of 4D along the temporal dimension is a 3D scan. However, as there are no previous works on 4D perception using neural networks, we will primarily cover 3D perception, specifically 3D segmentation using neural networks. We categorized all previous works in 3D as either (a) 3D-convolutional neural networks or (b) neural networks without 3D convolutions. Finally, we cover early 4D perception methods. Although 2D videos are spatio-temporal data, we will not cover them in this paper as 3D perception requires radically different data processing, implementation, and architectures.

\textbf{3D-convolutional neural networks.}
The first branch of 3D-convolutional neural networks uses a rectangular grid and a dense representation \cite{segcloud, scannet} where the empty space is represented either as $\mathbf{0}$ or the signed distance function. This straightforward representation is intuitive and is supported by all major public neural network libraries. However, as the most space in 3D scans is empty, it suffers from high memory consumption and slow computation. To resolve this, OctNet~\cite{octnet} proposed to use the Octree structure to represent 3D space and convolution on it.

The second branch is sparse 3D-convolutional neural networks~\cite{splatnet, sparse3dcnn}. There are two quantization methods used for high dimensions: a rectangular grid and a permutohedral lattice~\cite{permutohedral}. \cite{splatnet} used a permutohedral lattice whereas \cite{sparse3dcnn} used a rectangular grid for 3D classification and semantic segmentation.

The last branch is 3D pseudo-continuous convolutional neural networks~\cite{montecarlo, pointcnn}. Unlike the previous works, they define convolutions using continuous kernels in a continuous space. However, finding neighbors in a continuous space is expensive, as it requires KD-tree search rather than a hash table, and are susceptible to uneven distribution of point clouds.

\textbf{Neural networks without 3D convolutions.}
Recently, we saw a tremendous increase in neural networks without 3D convolutions for 3D perception. Since 3D scans consist of thin observable surfaces, \cite{surface, tangent} proposed to use 2D convolutions on the surface for semantic segmentation.

Another direction is PointNet-based methods~\cite{pointnet, pointnetpp}. PointNets use a set of input coordinates as features for a multi-layer perceptron. However, this approach processes a limited number of points and thus a sliding window for cropping out a section from an input was used for large spaces making the receptive field size rather limited. \cite{superpoint} tried to resolve such shortcomings with a recurrent network on top of multiple pointnets, and \cite{pointcnn} proposed a variant of 3D continuous convolution for lower layers of a PointNet and got a significant performance boost.

\textbf{4D perception.}
The first 4D perception algorithm~\cite{4dseg95} proposed a dynamic deformable balloon model for 4D cardiac image analysis. Later, \cite{4dmrf} used a 4D Markov Random Fields for cardiac segmentation. Recently, a "Spatio-Temporal CNN"~\cite{stcnn} combined a 3D-UNet with a 1D-AutoEncoder for temporal data and applied the model for auto-encoding brain fMRI images, but it is not a 4D-convolutional neural network.

In this paper, we propose the first high-dimensional convolutional neural networks for 4D spatio-temporal data, or 3D videos and the 7D space-time-chroma space. Compared with other approaches that combine temporal data with a recurrent neural network or a shallow model (CRF), our networks use a homogeneous representation and convolutions consistently throughout the networks. Instead of using an RNN, we use convolution for the temporal axis since it is proven to be more effective in sequence modeling~\cite{bai2018empirical}.

\section{Sparse Tensor and Convolution}
\label{sec:sparse}

In traditional speech, text, or image data, features are extracted densely. Thus, the most common representations of these data are vectors, matrices, and tensors. However, for 3-dimensional scans or even higher-dimensional spaces, such dense representations are inefficient due to the sparsity. Instead, we can only save the non-empty part of the space as its coordinates and the associated features. This representation is an N-dimensional extension of a sparse matrix; thus it is known as a sparse tensor. There are many ways to save such sparse tensors compactly~\cite{sparsetensor}, but we follow the COO format as it is efficient for neighborhood queries (Sec.~\ref{sec:sparse_convolution}). Unlike the traditional sparse tensors, we augment the sparse tensor coordinates with the batch indices to distinguish points that occupy the same coordinate in different batches~\cite{sparse3dcnn}. Concisely, we can represent a set of 4D coordinates as $\mathcal{C} = \{(x_i, y_i, z_i, t_i)\}_i$ or as a matrix $C$ and a set of associated features $\mathcal{F} = \{\mbf{f}_i\}_i$ or as a matrix $F$. Then, a sparse tensor can be written as
\begin{equation}
C = \begin{bmatrix}
x_1 & y_1 & z_1 & t_1 & b_1\\ 
    &     & \vdots &  &     \\ 
x_N & y_N & z_N & t_N & b_N
\end{bmatrix}, \;
F = \begin{bmatrix}
\mathbf{f}_1^T\\ 
\vdots\\ 
\mathbf{f}_N^T
\end{bmatrix}
\end{equation}
where $b_i$ and $\mathbf{f}_i$ are the batch index and the feature associated to the $i$-th coordinate. In Sec.~\ref{sec:crf}, we augment the 4D space with the 3D chromatic space and create a 7D sparse tensor for \textit{trilateral} filtering.

\subsection{Generalized Sparse Convolution}
\label{sec:sparse_convolution}

In this section, we generalize the sparse convolution~\cite{sparsecnn, sparse3dcnn} for generic input and output coordinates and for arbitrary kernel shapes. The generalized sparse convolution encompasses not only all sparse convolutions but also the conventional dense convolutions. Let $x^{\text{in}}_\mbf{u} \in \mathbb{R}^{N^\text{in}}$ be an $N^\text{in}$-dimensional input feature vector in a $D$-dimensional space at $\mbf{u} \in \mathbb{R}^D$ (a D-dimensional coordinate), and convolution kernel weights be $\mathbf{W} \in \mathbb{R}^{K^D
\times N^\text{out} \times N^\text{in}}$. We break down the weights into spatial weights with $K^D$ matrices of size $N^\text{out} \times N^\text{in}$ as $W_\mbf{i}$ for $|\{\mbf{i}\}_\mbf{i}| = K^D$. Then, the conventional dense convolution in D-dimension is
\begin{equation}
\mathbf{x}^{\text{out}}_\mbf{u} = \sum_{\mbf{i} \in \mathcal{V}^D(K)} W_\mbf{i} \mathbf{x}^{\text{in}}_{\mbf{u} + \mbf{i}} \text{ for } \mbf{u} \in \mathbb{Z}^D,
\label{eq:dense_convolution}
\end{equation}
where $\mathcal{V}^D(K)$ is the list of offsets in D-dimensional hypercube centered at the origin. e.g. $\mathcal{V}^1(3)=\{-1, 0, 1\}$. The generalized sparse convolution in Eq.~\ref{eq:sparse_convolution} relaxes Eq.~\ref{eq:dense_convolution}.
\begin{equation}
    \mathbf{x}^{\text{out}}_\mbf{u} = \sum_{\mbf{i} \in \mathcal{N}^D(\mbf{u}, \mathcal{C}^{\text{in}})} W_\mbf{i} \mathbf{x}^{\text{in}}_{\mbf{u} + \mbf{i}} \text{ for } \mbf{u} \in \mathcal{C}^{\text{out}}
\label{eq:sparse_convolution}
\end{equation}
where $\mathcal{N}^D$ is a set of offsets that define the shape of a kernel and $\mathcal{N}^D(\mbf{u}, \mathcal{C}^\text{in})= \{\mbf{i} | \mbf{u} + \mbf{i} \in \mathcal{C}^\text{in}, \mbf{i} \in \mathcal{N}^D \}$ as the set of offsets from the current center, $\mbf{u}$, that exist in $\mathcal{C}^\text{in}$. $\mathcal{C}^\text{in}$ and $\mathcal{C}^\text{out}$ are predefined input and output coordinates of sparse tensors. First, note that the input coordinates and output coordinates are not necessarily the same. Second, we define the shape of the convolution kernel arbitrarily with $\mathcal{N}^D$. This generalization encompasses many special cases such as the dilated convolution and typical hypercubic kernels. Another interesting special case is the "sparse submanifold convolution"~\cite{sparse3dcnn} when we set $\mathcal{C}^\text{out} = \mathcal{C}^\text{in}$ and $\mathcal{N}^D = \mathcal{V}^D(K)$. If we set $\mathcal{C}^\text{in} = \mathcal{C}^\text{out} = \mathbb{Z}^D$ and $\mathcal{N}^D = \mathcal{V}^D(K)$, the generalized sparse convolution becomes the conventional dense convolution (Eq.~\ref{eq:dense_convolution}). If we define the $\mathcal{C}^\text{in}$ and $\mathcal{C}^\text{out}$ as multiples of a natural number and $\mathcal{N}^D = \mathcal{V}^D(K)$, we have a strided dense convolution.

\section{Minkowski Engine}
\label{sec:minkowskiengine}

In this section, we propose an open-source auto-differentiation library for sparse tensors and the generalized sparse convolution (Sec.~\ref{sec:sparse}). As it is an extensive library with many functions, we will only cover essential forward-pass functions.

\subsection{Sparse Tensor Quantization}
The first step in the sparse convolutional neural network is the data processing to generate a sparse tensor, which converts an input into unique coordinates, associated features, and optionally labels when training for semantic segmentation. In Alg.~\ref{alg:sparse_quantization}, we list the GPU function for this process. When a dense label is given, it is important that we ignore voxels with more than one unique labels. This can be done by marking these voxels with \code{IGNORE\_LABEL}. First, we convert all coordinates into hash keys and find all unique hashkey-label pairs to remove collisions. Note that \code{SortByKey}, \code{UniqueByKey}, and \code{ReduceByKey} are all standard \code{Thrust} library functions~\cite{thrust}.
\begin{algorithm}[h]
  \caption{GPU Sparse Tensor Quantization}
  \label{alg:sparse_quantization}
\begin{algorithmic}
		\State Inputs: coordinates $C_p  \in \mathbb{R}^{N \times D}$, features $F_p \in \mathbb{R}^{N \times N_f}$, target labels $\mbf{l} \in \mathbb{Z}_{+}^N$, quantization step size $v_l$
		\State $C_p'\leftarrow $ floor($C_p$ / $v_l$)
		\State $\mbf{k} \leftarrow $ hash($C_p'$), $\mbf{i} \leftarrow$ sequence(N)
		\State $((\mbf{i}', \mbf{l}'), k') \leftarrow $ SortByKey($(\mbf{i}, \mbf{l})$, key=$\mbf{k}$)
		\State $(\mbf{i}'', (\mbf{k}'', \mbf{l}'')) \leftarrow $ UniqueByKey($\mbf{i}'$, key=$(\mbf{k}', \mbf{l}')$)
		\State $(\mbf{l}''', \mbf{i}''') \leftarrow $ ReduceByKey(($\mbf{l}'', \mbf{i}''$), key=$\mbf{k}''$, fn=$f$)
		\State \Return $C_p'[\mbf{i}''', :], F_p[\mbf{i}''', :], \mbf{l}'''$
\end{algorithmic}
\end{algorithm}
The reduction function $f((l_x, i_x), (l_y, i_y)) => (\code{IGNORE\_LABEL}, i_x)$ takes label-key pairs and returns the \code{IGNORE\_LABEL} since at least two label-key pairs in the same key means there is a label collision. A CPU-version works similarly except that all reduction and sorting are processed serially.

\subsection{Generalized Sparse Convolution}

The next step in the pipeline is generating the output coordinates $\mathcal{C}^\text{out}$ given the input coordinates $\mathcal{C}^\text{in}$ (Eq.~\ref{eq:sparse_convolution}). When used in conventional neural networks, this process requires only a convolution stride size, input coordinates, and the stride size of the input sparse tensor (the minimum distance between coordinates). The algorithm is presented in the supplementary material. We create this output coordinates dynamically allowing an arbitrary output coordinates $\mathcal{C}^\text{out}$ for the generalized sparse convolution.

Next, to convolve an input with a kernel, we need a mapping to identify which inputs affect which outputs. This mapping is not required in conventional dense convolutions as it can be inferred easily. However, for sparse convolution where coordinates are scattered arbitrarily, we need to specify the mapping. We call this mapping the kernel maps and define them as pairs of lists of input indices and output indices, $\mbf{M} = \{(I_\mbf{i}, O_\mbf{i})\}_\mbf{i}$ for $\mbf{i} \in \mathcal{N}^D$. Finally, given the input and output coordinates, the kernel map, and the kernel weights $W_\mbf{i}$, we can compute the generalized sparse convolution by iterating through each of the offset $\mbf{i} \in \mathcal{N}^D$ (Alg.~\ref{alg:sparse_convolution})
\begin{algorithm}[h]
\begin{algorithmic}[1]
\Require Kernel weights $\mbf{W}$, input features $F^i$, output feature placeholder $F^o$, convolution mapping $\mbf{M}$, 
\State $F^o \leftarrow \mbf{0}$  // set to 0
\ForAll{$W_\mbf{i}, (I_\mbf{i}, O_\mbf{i}) \in (\mbf{W}, \mbf{M})$}
    \State $F_{\text{tmp}} \leftarrow W_\mbf{i} [F^i_{I_\mbf{i}[1]}, F^i_{I_\mbf{i}[2]}, ...,  F^i_{I_\mbf{i}[n]}]$ // (cu)BLAS
    \State $F_{\text{tmp}} \leftarrow F_{\text{tmp}} + [F^o_{O_\mbf{i}[1]}, F^o_{O_\mbf{i}[2]}, ...,  F^o_{O_\mbf{i}[n]}]$
    \State $[F^o_{O_\mbf{i}[1]}, F^o_{O_\mbf{i}[2]}, ...,  F^o_{O_\mbf{i}[n]}] \leftarrow F_{\text{tmp}}$
\EndFor
\end{algorithmic}
\caption{Generalized Sparse Convolution}
\label{alg:sparse_convolution}
\end{algorithm}
where $I[n]$ and $O[n]$ indicate the $n$-th element of the list of indices $I$ and $O$ respectively and $F^i_n$ and $F^o_n$ are also $n$-th input and output feature vectors respectively. The transposed generalized sparse convolution (deconvolution) works similarly except that the role of input and output coordinates is reversed.

\subsection{Max Pooling}

Unlike dense tensors, on sparse tensors, the number of input features varies per output. Thus, this creates non-trivial implementation for a max/average pooling. Let $\mbf{I}$ and $\mbf{O}$ be the vector that concatenated all $\{I_\mbf{i}\}_\mbf{i}$ and $\{O_\mbf{i}\}_\mbf{i}$ for $\mbf{i} \in \mathcal{N}^D$ respectively. We first find the number of inputs per each output coordinate and indices of the those inputs. Alg.~\ref{alg:max_pooling} reduces the input features that map to the same output coordinate. \code{Sequence}(n) generates a sequence of integers from 0 to n - 1 and the reduction function $f((k_1, v_1), (k_2, v_2)) = \min(v_1, v_2)$ which returns the minimum value given two key-value pairs. \code{MaxPoolKernel} is a custom CUDA kernel that reduces all features at a specified channel using $\mbf{S}'$, which contains the first index of $\mbf{I}$ that maps to the same output, and the corresponding output indices $\mbf{O}"$.
\begin{algorithm}
  \caption{GPU Sparse Tensor MaxPooling}
  \label{alg:max_pooling}
\begin{algorithmic}
  \State Input: input feature $F$, output mapping $\mbf{O}$
  \State $(\mbf{I}', \mbf{O}') \leftarrow$ SortByKey($\mbf{I}$, key=$\mbf{O}$)
  \State $\mbf{S} \leftarrow$ Sequence(length($\mbf{O}'$))
  \State $\mbf{S}', \mbf{O}" \leftarrow$ ReduceByKey($\mbf{S}$, key=$\mbf{O}'$, fn=$f$)
  \State \Return MaxPoolKernel($\mbf{S}'$, $\mbf{I}'$, $\mbf{O}"$, $F$)
\end{algorithmic}
\end{algorithm}

\subsection{Global / Average Pooling, Sum Pooling}

An average pooling and a global pooling layer compute the average of input features for each output coordinate for average pooling or one output coordinate for global pooling. This can be implemented in multiple ways. We use a sparse matrix multiplication since it can be optimized on hardware or using a faster sparse BLAS library. In particular, we use the \codeword{cuSparse} library for sparse matrix-matrix (\codeword{cusparse_csrmm}) and matrix-vector multiplication (\codeword{cusparse_csrmv}) to implement these layers. Similar to the max pooling algorithm, $\mbf{M}$ is the $(\mbf{I}, \mbf{O})$ input-to-output kernel map. For the global pooling, we create the kernel map that maps all inputs to the origin and use the same Alg.~\ref{alg:avg_pooling}. The transposed pooling (unpooling) works similarly.

On the last line of the Alg.~\ref{alg:avg_pooling}, we divide the pooled features by the number of inputs mapped to each output. However, this process could remove density information. Thus, we propose a variation that does not divide the number of inputs and named it the sum pooling.
\begin{algorithm}
  \caption{GPU Sparse Tensor AvgPooling}
  \label{alg:avg_pooling}
\begin{algorithmic}
  \State Input: mapping $\mbf{M} = (\mbf{I}, \mbf{O})$, features $F$, one vector $\mbf{1}$
  \State $S_M$ = coo2csr(row=$\mbf{O}$, col=$\mbf{I}$, val=$\mbf{1}$)
  \State $F'$ = cusparse\_csrmm($S_M$, $F$)
  \State $N$ = cusparse\_csrmv($S_M$, $\mbf{1}$)
  \State \Return $F' / N$
\end{algorithmic}
\end{algorithm}

\subsection{Non-spatial Functions}
For functions that do not require spatial information (coordinates) such as ReLU, we can apply the functions directly to the features $F$. Also, for batch normalization, as each row of $F$ represents a feature, we could use the 1D batch normalization function directly on $F$.

\section{Minkowski Convolutional Neural Networks}
\label{sec:minkowski}

In this section, we introduce 4-dimensional spatio-temporal convolutional neural networks for spatio-temporal perception. We treat the time dimension as an extra spatial dimension and create networks with 4-dimensional convolutions. However, there are unique problems arising from high-dimensional convolutions. First, the computational cost and the number of parameters in the networks increase exponentially as we increase the dimension. However, we experimentally show that these increases do not necessarily lead to better performance. Second, the networks do not have an incentive to make the prediction consistent throughout the space and time with conventional cross-entropy loss alone.

To resolve the first problem, we make use of a special property of the generalized sparse convolution and propose non-conventional kernel shapes that not only save memory and computation, but also perform better. Second, to enforce spatio-temporal consistency, we propose a high-dimensional conditional random field (7D space-time-color space) that filters network predictions. We use variational inference to train both the base network and the conditional random field end-to-end.

\subsection{Tesseract Kernel and Hybrid Kernel}
\label{sec:hybrid_kernel}

The surface area of 3D data increases linearly to time and quadratically to the spatial resolution. However, when we use a conventional 4D hypercube, or a tesseract (Fig.~\ref{fig:hypercubes}), for a convolution kernel, the exponential increase in the number of parameters leads to over-parametrization, overfitting, as well as high computational-cost and memory consumption. Instead, we propose a hybrid kernel (non-hypercubic, non-permutohedral) to save computation. We use the arbitrary kernel offsets $\mathcal{N}^D$ of the generalized sparse convolution to implement the hybrid kernel.

\begin{figure}[ht]
    \centering
    \minipage{0.2\columnwidth}
      \includegraphics[width=\linewidth]{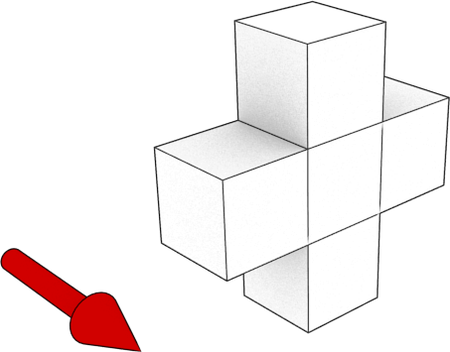}
      \caption*{\small{Cross}}\label{fig:cross_kernel}
    \endminipage
    \minipage{0.2\columnwidth}
      \includegraphics[width=\linewidth]{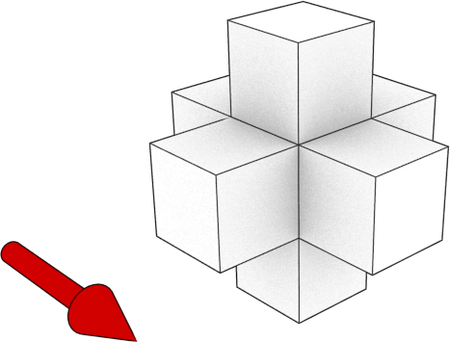}
      \caption*{\small{Hypercross}}\label{fig:hypercross_kernel}
    \endminipage
    \minipage{0.2\columnwidth}
        \includegraphics[width=0.95\linewidth]{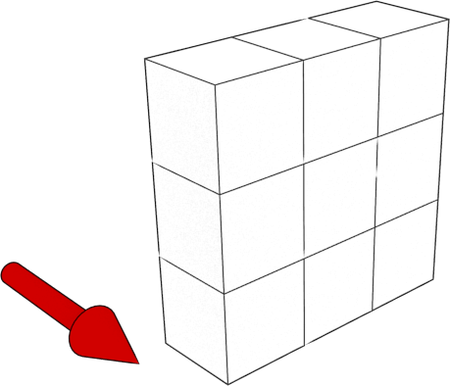}
      \caption*{\small{Cube}}\label{fig:cube_kernel}
    \endminipage\hfill
    \minipage{0.2\columnwidth}
        \includegraphics[width=0.95\linewidth]{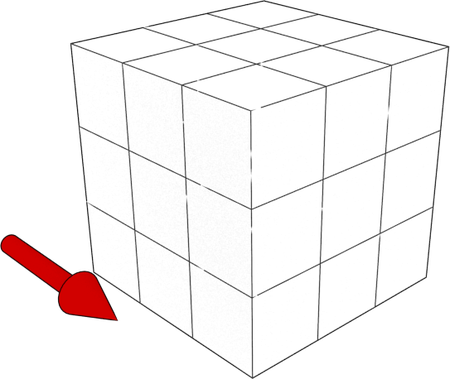}
      \caption*{\small{Hypercube}}\label{fig:hypercube_kernel}
    \endminipage\hfill
    \minipage{0.2\columnwidth}
        \includegraphics[width=0.95\linewidth]{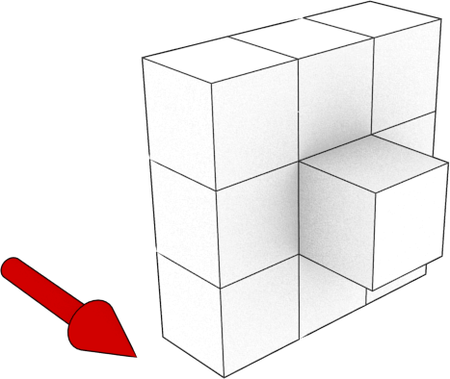}
      \caption*{\small{Hybrid}}\label{fig:hybrid_kernel}
    \endminipage\hfill
    \caption{Various kernels in space-time. The red arrow indicates the temporal dimension and the other two axes are for spatial dimensions. The third spatial dimension is hidden for better visualization.}
    \label{fig:kernels}
    \vspace{-1em}
\end{figure}

The hybrid kernel is a combination of a cross-shaped kernel a conventional cubic kernel (Fig.~\ref{fig:kernels}). For spatial dimensions, we use a cubic kernel to capture the spatial geometry accurately. For the temporal dimension, we use the cross-shaped kernel to connect the same point in space across time. We experimentally show that the hybrid kernel outperforms the tesseract kernel both in speed and accuracy.

\begin{figure}
  \centering
    \includegraphics[width=0.48\columnwidth]{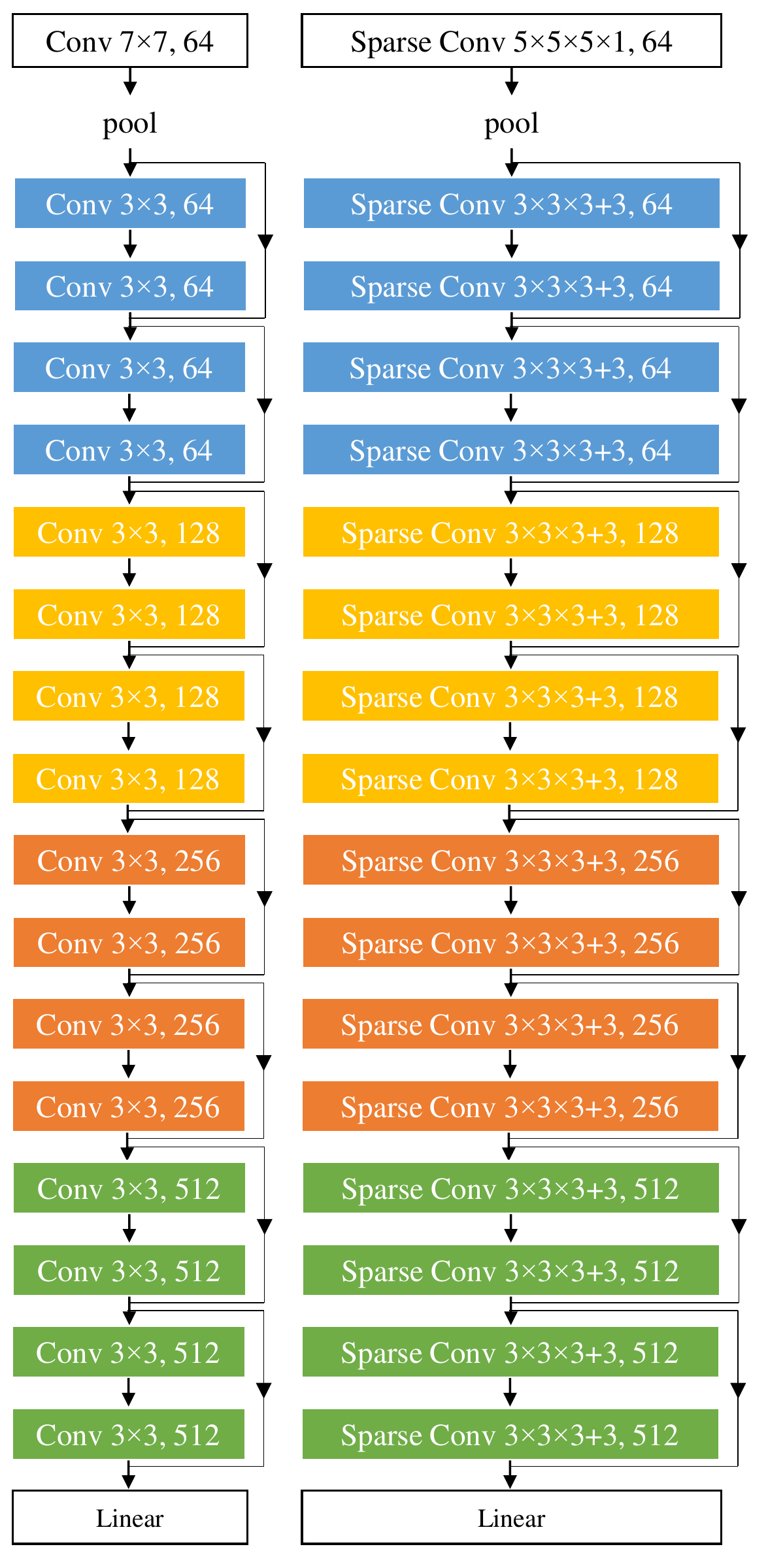}
		\vspace{-0.5em}
		\caption{\small{Architecture of ResNet18 (left) and MinkowskiNet18 (right). Note the structural similarity. $\times$ indicates a hypercubic kernel, $+$ indicates a hypercross kernel. (best viewed on display)}}
  \label{fig:resnet}
  \vspace{-1.7em}
\end{figure}

\begin{figure*}
  \centering
    \includegraphics[width=0.7\textwidth]{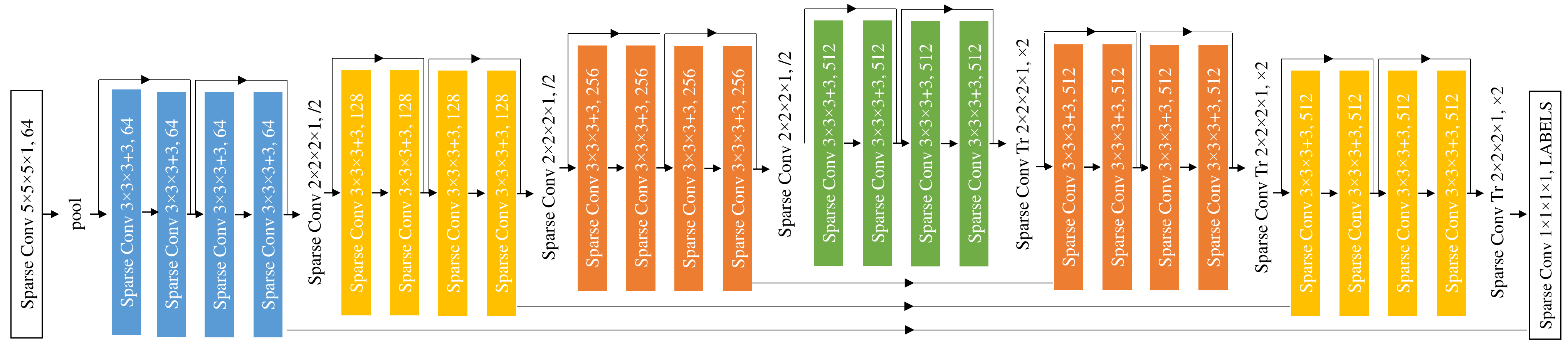}
		\vspace{-0.5em}
		\caption{\small{Architecture of MinkowskiUNet32. $\times$ indicates a hypercubic kernel, $+$ indicates a hypercross kernel. (best viewed on display)}}
  \label{fig:resunet}
  \vspace{-1.5em}
\end{figure*}

\subsection{Residual Minkowski Networks}
The generalized sparse convolution allows us to define strides and kernel shapes arbitrarily. Thus, we can create a high-dimensional network only with generalized sparse convolutions, making the implementation easier and generic. In addition, it allows us to adopt recent architectural innovations in 2D directly to high-dimensional networks. To demonstrate, we create a high-dimensional version of a residual network on Fig.~\ref{fig:resnet}. For the first layer, instead of a $7 \times 7$ 2D convolution, we use a $5\times 5\times 5 \times 1$ generalized sparse convolution. However, for the rest of the networks, we follow the original network architecture.

For the U-shaped variants, we add multiple strided sparse convolutions and strided sparse transpose convolutions with skip connections connecting the layers with the same stride size (Fig.~\ref{fig:resunet}) on the base residual network. We use multiple variations of the same architecture for semantic segmentation experiments.

\section{Trilateral Stationary-CRF}
\label{sec:crf}

For semantic segmentation, the cross-entropy loss is applied for each pixel or voxel. However, the loss does not enforce consistency as it does not have pair-wise terms. To make such consistency more explicit, we propose a high-dimensional conditional random field (CRF) similar to the one used in image semantic segmentation~\cite{crfasrnn}. In image segmentation, the bilateral space that consists of 2D space and 3D color is used for the CRF. For 3D-videos, we use the trilateral space that consists of 3D space, 1D time, and 3D chromatic space. The color space creates a "spatial" gap between points with different colors that are spatially adjacent (e.g., on a boundary). Thus, it prevents information from "leaking out" to different regions. Unlike conventional CRFs with Gaussian edge potentials and dense connections~\cite{densecrf,crfasrnn}, we do not restrict the compatibility function to be a Gaussian. Instead, we relax the constraint and only apply the stationarity condition.

To find the global optima of the distribution, we use the variational inference and convert a series of fixed point update equations to a recurrent neural network similar to~\cite{crfasrnn}. We use the generalized sparse convolution in 7D space to implement the recurrence and jointly train both the base network that generates unary potentials and the CRF end-to-end.

\subsection{Definition}

Let a CRF node in the 7D (space-time-chroma) space be $x_i$ and the unary potential be $\phi_u(x_i)$ and the pairwise potential as $\phi_p(x_i, x_j)$ where $x_j$ is a neighbor of $x_i$, $\mathcal{N}^7(x_i)$. The conditional random field is defined as
\begin{equation*}
    P(\mbf{X}) = \frac{1}{Z} \exp{ \sum_i \left( \phi_u(x_i) + \sum_{j \in \mathcal{N}^7(x_i)} \phi_p(x_i, x_j)\right)}
\end{equation*}
where $Z$ is the partition function; $X$ is the set of all nodes; and $\phi_p$ must satisfy the \textit{stationarity} condition $\phi_p(\mbf{u}, \mbf{v}) = \phi_p(\mbf{u} + \mbf{\tau_u}, \mbf{v} + \mbf{\tau_v})$ for $\mbf{\tau_u}, \mbf{\tau_v} \in \mathbb{R}^D$. Note that we use the camera extrinsics to define the spatial coordinates of a node $x_i$ in the world coordinate system. This allows stationary points to have the same coordinates throughout the time.

\subsection{Variational Inference}
The optimization $\argmax_\mbf{X} P(X)$ is intractable. Instead, we use the variational inference to minimize divergence between the optimal $P(\mbf{X})$ and an approximated distribution $Q(\mbf{X})$.
Specifically, we use the mean-field approximation, $Q = \prod_i Q_i(x_i)$ as the closed form solution exists. From the Theorem 11.9 in~\cite{pgm}, $Q$ is a local maximum if and only if
\begin{equation*}
    Q_i(x_i) = \frac{1}{Z_i}\exp \underset{\mbf{X}_{-i} \sim Q_{-i}}{\mathbf{E}} \left[\phi_u(x_i) + \sum_{j \in \mathcal{N}^7(x_
    i)} \phi_p(x_i, x_j) \right].
\end{equation*}
$\mbf{X}_{-i}$ and $Q_{-i}$ indicate all nodes or variables except for the $i$-th one. The final fixed-point equation is Eq.~\ref{eq:vi_update}. The derivation is in the supplementary material.
\begin{equation} \label{eq:vi_update}
		Q_i^+(x_i) = \frac{1}{Z_i}\exp \left\{ \phi_u(x_i) + \sum_{j \in \mathcal{N}^7(x_i)} \sum_{x_j} \phi_p(x_i, x_j)Q_j(x_j) \right\}
\end{equation}

\subsection{Learning with 7D Sparse Convolution}
Interestingly, the weighted sum $\phi_p(x_i, x_j)Q_j(x_j)$ in Eq.~\ref{eq:vi_update} is equivalent to a generalized sparse convolution in the 7D space since $\phi_p$ is \textit{stationary} and each edge between $x_i, x_j$ can be encoded using $\mathcal{N}^7$. Thus, we convert fixed point update equation Eq.~\ref{eq:vi_update} into an algorithm in Alg.~\ref{alg:crfvi}.
\begin{algorithm}[h]
  \caption{Variational Inference of TS-CRF}
  \label{alg:crfvi}
\begin{algorithmic}
  \Require Input: Logit scores $\phi_u$ for all $x_i$; associated coordinate $C_i$, color $F_i$, time $T_i$
    \State $Q^0(X) = \exp{\phi_u(X)}$, $C_\text{crf} = [C, F, T]$
    \For{$n$ from 1 to $N$}
      \State $\tilde{Q}^n = $ SparseConvolution($(C_\text{crf}, Q^{n-1})$, kernel=$\phi_p$)
      \State $Q^n = $ Softmax($\phi_u + \tilde{Q}^n$)
    \EndFor
    \State \Return $Q^N$
\end{algorithmic}
\end{algorithm}

Finally, we use $\phi_u$ as the logit predictions of a 4D Minkowski network and train both $\phi_u$ and $\phi_p$ \textit{end-to-end} using one 4D and one 7D Minkowski network using Eq.~\ref{eq:crfasrnn}.
\begin{equation} \label{eq:crfasrnn}
    \frac{\partial L}{\partial \phi_p} = \sum_n^N \frac{\partial L}{\partial Q^{n+}} \frac{\partial Q^{n+}}{\partial \phi_p}, \;\;
    \frac{\partial L}{\partial \phi_u} = \sum_n^N \frac{\partial L}{\partial Q^{n+}} \frac{\partial Q^{n+}}{\partial \phi_u}
\end{equation}

\section{Experiments}
\label{sec:experiments}

To validate the proposed high-dimensional networks, we first use multiple standard 3D benchmarks for 3D semantic segmentation. It allows us to gauge the performance of the high-dimensional networks with the same architecture with other state-of-the-art methods. Next, we create multiple 4D datasets from 3D datasets that have temporal sequences and analyze each of the proposed components for ablation study.

\subsection{Implementation}
We implemented the Minkowski Engine (Sec.~\ref{sec:minkowskiengine}) using C++/CUDA and wrap it with PyTorch~\cite{pytorch}. Data is prepared in parallel data processes that load point clouds, apply data augmentation, and quantize them with Alg.~\ref{alg:sparse_quantization} on the fly. For non-spatial functions, we use the PyTorch functions directly.

\subsection{Training and Evaluation}
We use Momentum SGD with the Poly scheduler to train networks from learning rate $1\text{e-}1$ and apply data augmentation including random scaling, rotation around the gravity axis, spatial translation, spatial elastic distortion, and chromatic translation and jitter.

For evaluation, we use the standard mean Intersection over Union (mIoU) and mean Accuracy (mAcc) for metrics following the previous works. To convert voxel-level predictions to point-level predictions, we simply propagated predictions from the nearest voxel center.

\subsection{Datasets}

\textbf{ScanNet.}
The ScanNet~\cite{scannet} 3D segmentation benchmark consists of 3D reconstructions of real rooms. It contains ~1.5k rooms, some repeated rooms captured with different sensors. We feed an entire room to a MinkowskiNet fully convolutionally without cropping.

\textbf{Stanford 3D Indoor Spaces (S3DIS).}
The dataset~\cite{stanford3dis} contains 3D scans of six floors of three different buildings. We use the Fold \#1 split following many previous works. We use 5cm and 2cm voxel for the experiment.

\textbf{RueMonge 2014 (Varcity).} The RueMonge 2014 dataset~\cite{ruemonge2014} provides semantic labels for a multi-view 3D reconstruction of the Rue Mongue. To create a 4D dataset, we crop the 3D reconstruction on-the-fly to generate a temporal sequence. We use the official split for all experiments.

\textbf{Synthia 4D.} We use the Synthia dataset~\cite{synthia} to create 3D video sequences. We use 6 sequences of driving scenarios in 9 different weather conditions. Each sequence consists of 4 stereo RGB-D images taken from the top of a car. We back-project the depth images to the 3D space to create 3D videos. We visualized a part of a sequence in Fig.~\ref{fig:3dvideo}.

We use the sequence 1-4 except for sunset, spring, and fog for the train split; the sequence 5 foggy weather for validation; and the sequence 6 sunset and spring for the test. In total, the train/val/test set contain 20k/815/1886 3D scenes respectively.

Since the dataset is purely synthetic, we added various noise to the input point clouds to simulate noisy observations. We used elastic distortion, Gaussian noise, and chromatic shift in the color for the noisy 4D Synthia experiments.

\subsection{Results and Analysis}

{\small
\tabcolsep= 1mm
\begin{table}[t]
    \centering
    \caption{3D Semantic Label Benchmark on ScanNet${}^\dagger$~\cite{scannet}}
    \resizebox{0.5\columnwidth}{!}{
    \begin{tabular}{r|c}
    \toprule
            \small{Method}  & \small{mIOU} \\
    \midrule
        ScanNet~\cite{scannet}  & 30.6\\
        SSC-UNet~\cite{sparse3dsegmentation} & 30.8 \\
        PointNet++~\cite{pointnetpp} & 33.9\\
        ScanNet-FTSDF & 38.3 \\
        SPLATNet~\cite{splatnet} & 39.3 \\
        TangetConv~\cite{tangent} & 43.8 \\
        SurfaceConv \cite{surface} & 44.2 \\
        3DMV${}^\ddagger$~\cite{3dmv} & 48.4 \\
        3DMV-FTSDF${}^\ddagger$ & 50.1 \\
        PointNet++SW & 52.3 \\
    \midrule
        MinkowskiNet42 (5cm) & \textbf{67.9} \\
    \midrule
        SparseConvNet~\cite{sparse3dsegmentation}${}^\dagger$ & 72.5 \\
        MinkowskiNet42 (2cm)${}^\dagger$ & \textbf{73.4} \\
    \bottomrule
    \end{tabular}}
    \label{tab:scannet}
    \caption*{\small{${}^\dagger$: post-CVPR submissions. ${}^\ddagger$: uses 2D images additionally. Per class IoU in the supplementary material. The parenthesis next to our methods indicate the voxel size.}}
\end{table}}

{\small
\tabcolsep= 1mm
\begin{table}[ht]
    \centering
    \caption{Segmentation results on the 4D Synthia dataset}
    \resizebox{0.65\columnwidth}{!}{
    \begin{tabular}{l|cc}
    \toprule
        \small{Method}  & \small{mIOU} & \small{mAcc} \\
    \midrule
        3D MinkNet20  & 76.24 & 89.31 \\
        3D MinkNet20 + TA & 77.03 & 89.20 \\
    \midrule
        4D Tesseract MinkNet20 & 75.34 & 89.27 \\
        4D MinkNet20  & 77.46 & 88.013\\
        4D MinkNet20 + TS-CRF & 78.30 & 90.23 \\
        4D MinkNet32 + TS-CRF & \textbf{78.67} & \textbf{90.51} \\
    \bottomrule
    \end{tabular}}
    \caption*{\small{TA denotes temporal averaging. Per class IoU in the supplementary material.}}
    \label{tab:synthia}
\end{table}}

\begin{table*}[ht]
    \centering
    \caption{Segmentation results on the noisy Synthia 4D dataset}
    \resizebox{0.99\textwidth}{!}{
    \begin{tabular}{l|ccccccccccc|c}
    \toprule
        \small{IoU}  & \small{Building} & \small{Road} & \small{Sidewalk} & \small{Fence} & \small{Vegetation} & \small{Pole} & \small{Car} & \small{Traffic Sign} & \small{Pedestrian} & \small{Lanemarking} & \small{Traffic Light} & \small{mIoU} \\
    \midrule
        3D MinkNet42               & 87.954              & 97.511          & 78.346          & 84.307          & 96.225          & 94.785          & 87.370          & 42.705          & \textbf{66.666} & 52.665          & 55.353          & 76.717 \\
        3D MinkNet42 + TA          & 87.796              & 97.068          & 78.500          & 83.938          & 96.290          & 94.764          & 85.248          & 43.723          & 62.048          & 50.319          & 54.825          & 75.865 \\
    \midrule
	    4D Tesseract MinkNet42 & \textbf{89.957}     & 96.917          & 81.755          & 82.841          & 96.556          & 96.042          & 91.196          & 52.149          & 51.824          & \textbf{70.388} & \textbf{57.960} & 78.871 \\
	    4D MinkNet42 &                        88.890 & \textbf{97.720} & \textbf{85.206} & \textbf{84.855} & \textbf{97.325} & \textbf{96.147} & \textbf{92.209} & \textbf{61.794} & 61.647          & 55.673          & 56.735          & \textbf{79.836} \\
    \bottomrule
    \end{tabular}}
    \caption*{\small{TA denotes temporal averaging. As the input pointcloud coordinates are noisy, averaging along the temporal dimension introduces noise.}}
    \label{tab:noisy_synthia}
\end{table*}

\begin{figure}[t]
  \centering
  \begin{tabular}{c|c}
    \hspace{-1em}
    \includegraphics[width=0.45\columnwidth]{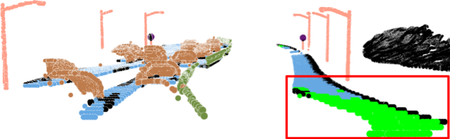} &
    \includegraphics[width=0.45\columnwidth]{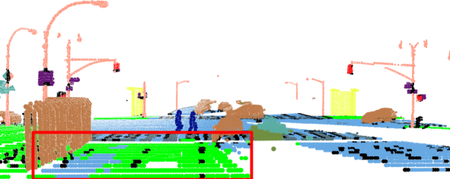} \\
    \includegraphics[width=0.45\columnwidth]{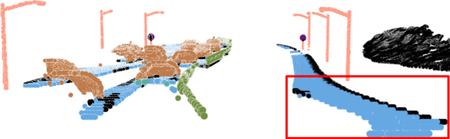} &
    \includegraphics[width=0.45\columnwidth]{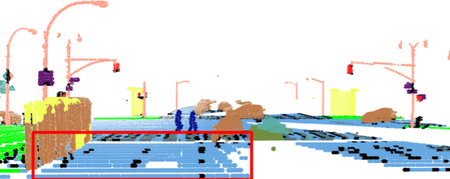} \\
  \end{tabular}
		\caption{\small{Visualizations of 3D (top), and 4D networks (bottom) on Synthia. A road (blue) far away from the car is often confused as sidewalks (green) with a 3D network, which persists after temporal averaging. However, 4D networks accurately captured it.}}
  \label{fig:synthia}
\end{figure}

\newcommand\scannetfigwidth{0.48\columnwidth}
\begin{figure}[t]
\centering
\begin{tabular}{c|c}
\includegraphics[width=\scannetfigwidth]{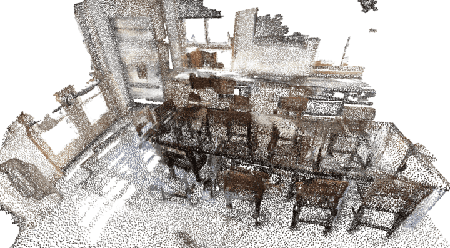} & \includegraphics[width=\scannetfigwidth]{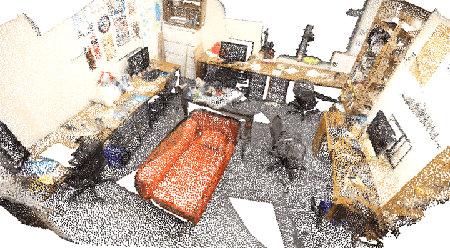} \\
\includegraphics[width=\scannetfigwidth]{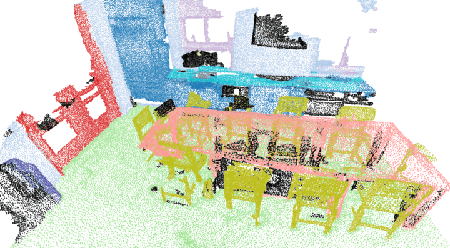} & \includegraphics[width=\scannetfigwidth]{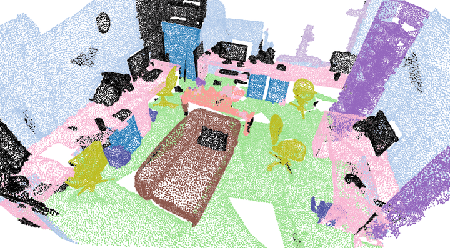} \\
\includegraphics[width=\scannetfigwidth]{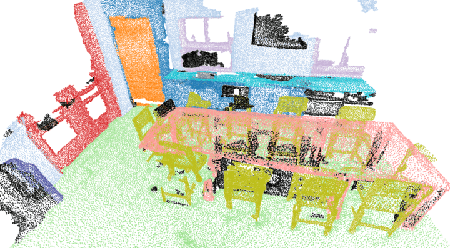} &
\includegraphics[width=\scannetfigwidth]{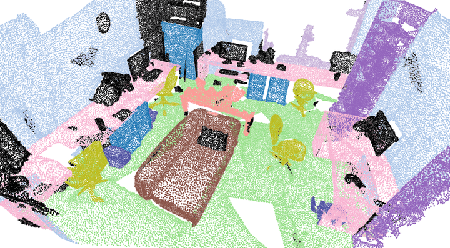} \\
\end{tabular}
\caption{\small{Visualization of Scannet predictions. From the top, a 3D input pointcloud, a network prediction, and the ground-truth.}}
\end{figure}

{\tabcolsep= 1mm
\begin{table}[t]
    \small
    \centering
    \caption{Stanford Area 5 Test (Fold \#1) (S3DIS)~\cite{stanford3dis}}
    \resizebox{0.58\columnwidth}{!}{
    \begin{tabular}{r|cc}
    \toprule
        Method & mIOU & mAcc\\
    \midrule
        PointNet \cite{pointnet} & 41.09 & 48.98\\
        SparseUNet \cite{sparse3dcnn} & 41.72 & 64.62 \\
        SegCloud \cite{segcloud} & 48.92 & 57.35 \\
        TangentConv \cite{tangent}  & 52.8 & 60.7 \\
        3D RNN \cite{3drnn} & 53.4 & 71.3 \\
        PointCNN \cite{pointcnn}  & 57.26 & 63.86 \\
        SuperpointGraph \cite{superpoint} & 58.04 & 66.5 \\
    \midrule
        MinkowskiNet20 & 62.60 & 69.62 \\ 
			MinkowskiNet32 & \textbf{65.35} & \textbf{71.71} \\
    \bottomrule
    \end{tabular}
    }
    \caption*{\small{Per class IoU in the supplementary material.}}
    \label{tab:S3DIS}
\end{table}}

\newcommand\stanfordfigwidth{0.48\columnwidth}
\begin{figure}[h]
\centering
\begin{tabular}{c|c}
\includegraphics[width=\stanfordfigwidth]{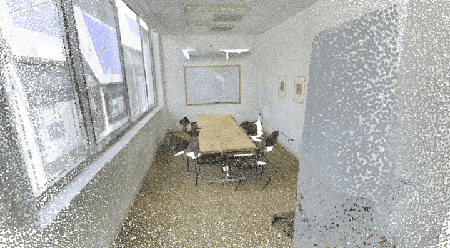} & \includegraphics[width=\stanfordfigwidth]{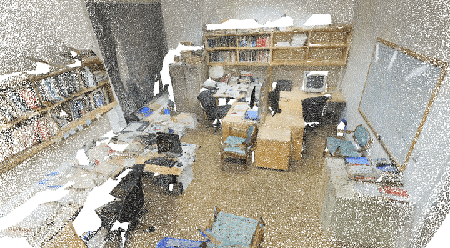} \\
\includegraphics[width=\stanfordfigwidth]{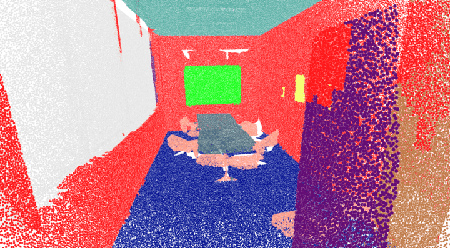} & \includegraphics[width=\stanfordfigwidth]{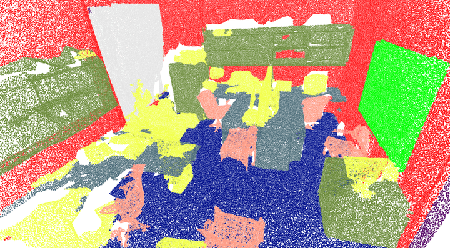} \\
\includegraphics[width=\stanfordfigwidth]{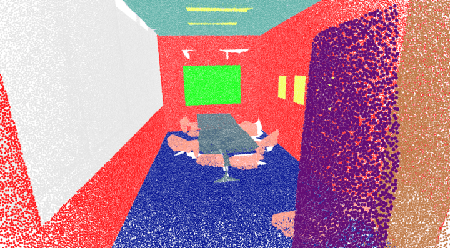} &
\includegraphics[width=\stanfordfigwidth]{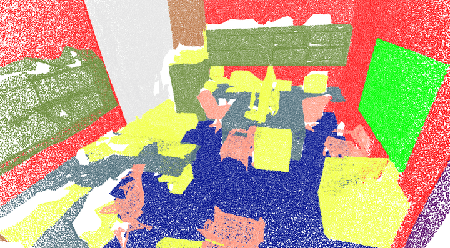} \\
\end{tabular}
\caption{\small{Visualization of Stanford dataset Area 5 test results. From the top, RGB input, prediction, ground truth.}}
\end{figure}

{\small
\tabcolsep= 1mm
\begin{table}
    \centering
    \caption{RueMonge 2014 dataset (Varcity) TASK3~\cite{ruemonge2014}}
    \resizebox{0.54\columnwidth}{!}{
    \begin{tabular}{l|c}
    \toprule
        \small{Method}  & \small{mIOU} \\
    \midrule
        MV-CRF~\cite{riemenschneider2014learning} & 42.3 \\
        Gradde et al.~\cite{gadde2017efficient} & 54.4 \\
        RF+3D CRF~\cite{martinovic20153d} & 56.4 \\
        OctNet ($256^3$)~\cite{octnet} & 59.2\\
        SPLATNet (3D) \cite{splatnet} & 65.4 \\
    \midrule
        3D MinkNet20 & 66.46 \\ 
        4D MinkNet20 & 66.56 \\ 
        4D MinkNet20 + TS-CRF & \textbf{66.59} \\ 
        
    \bottomrule
    \end{tabular}}
    \caption*{\small{The performance saturates quickly due to the small training set. Per class IoU in the supplementary material.}}
    \label{tab:varcity}
\end{table}}

\textbf{ScanNet \& Stanford 3D Indoor} The ScanNet and the Stanford Indoor datasets are one of the largest non-synthetic datasets, which make the datasets ideal test beds for 3D segmentation. We were able to achieve +19\% mIOU on ScanNet, and +7\% on Stanford compared with the best-published works by the CVPR deadline. This is due to the depth of the networks and the fine resolution of the space. We trained the same network for 60k iterations with 2cm voxel and achieved 72.1\% mIoU on ScanNet after the deadline. For all evaluation, we feed an entire room to a network and process it fully convolutionally.

\textbf{4D analysis} The RueMongue dataset is a small dataset that ranges one section of a street, so with the smallest network, we were able to achieve the best result (Tab.~\ref{tab:varcity}). However, the results quickly saturate. On the other hand, the Synthia 4D dataset has an order of magnitude more 3D scans than any other datasets, so it is more suitable for the ablation study.

We use the Synthia datasets with and without noise for 3D and 4D analysis and results are presented in Tab.~\ref{tab:synthia} and Tab.~\ref{tab:noisy_synthia}. We use various 3D and 4D networks with and without TS-CRF. Specifically, when we simulate noise in sensory inputs on the 4D Synthia dataset, we can observe that the 4D networks are more robust to noise. Note that the number of parameters added to the 4D network compared with the 3D network is less than 6.4 \% and $6\text{e-}3$ \% for the TS-CRF. Thus, with a small increase in computation, we could get a more robust algorithm with higher accuracy. In addition, when we process temporal sequence using the 4D networks, we could even get small speed gain as we process data in a batch mode. On Tab.~\ref{tab:runtime}, we vary the voxel size and the sequence length and measured the runtime of the 3D and 4D networks, as well as the 4D networks with TS-CRF. Note that for large voxel sizes, we tend to get small speed gain on the 4D networks compared with 3D networks.



{\small
\tabcolsep= 1mm
\begin{table}[h]
    \centering
    \caption{\small{Time (s) to process 3D videos with 3D and 4D MinkNet, the volume of a scan at each time step is 50m$\times$ 50m $\times$ 50m}}
    \resizebox{0.95\columnwidth}{!}{
    \begin{tabular}{c|ccc|ccc|ccc}
    \toprule
\small{Voxel Size}   & \multicolumn{3}{c}{0.6m} & \multicolumn{3}{c}{0.45m} & \multicolumn{3}{c}{0.3m} \\
    \midrule
\small{Video Length (s)} & 3D          & 4D         & 4D-CRF & 3D          & 4D          & 4D-CRF & 3D          & 4D         & 4D-CRF \\
    \midrule
3            & 0.18        & 0.14       & 0.17   & 0.25        & 0.22    &  0.27      & 0.43        & 0.49    & 0.59   \\
5            & 0.31        & 0.23       & 0.27   & 0.41        & 0.39    &  0.47      & 0.71        & 0.94    & 1.13   \\
7            & 0.43        & 0.31       & 0.38   & 0.58        & 0.61    &  0.74      & 0.99        & 1.59    & 2.02   \\
\bottomrule
\end{tabular}
}
    \label{tab:runtime}
\end{table}}

\section{Conclusion}

In this paper, we propose a generalized sparse convolution and an auto-differentiation library for sparse tensors and the generalized sparse convolution. Using these, we create 4D convolutional neural networks for spatio-temporal perception. Experimentally, we show that 3D convolutional neural networks alone can outperform 2D networks and 4D perception can be more robust to noise.

\section{Acknowledgements}

Toyota Research Institute ("TRI") provided funds to assist the authors with their research but this article solely reflects the opinions and conclusions of its authors and not TRI or any other Toyota entity. We acknowledge the support of the System X Fellowship and the companies sponsored: NEC Corporation, Nvidia, Samsung, and Tencent. Also, we want to acknowledge the academic hardware donation from Nvidia.

{
\begingroup
\let\clearpage\relax
    {\small
    \bibliographystyle{ieee_fullname}
    \bibliography{egbib}

\begin{thebibliography}{10}\itemsep=-1pt

\bibitem{permutohedral}
Andrew Adams, Jongmin Baek, and Myers~Abraham Davis.
\newblock Fast high-dimensional filtering using the permutohedral lattice.
\newblock In {\em Computer Graphics Forum}, volume~29, pages 753--762. Wiley
  Online Library, 2010.

\bibitem{stanford3dis}
Iro Armeni, Ozan Sener, Amir~R. Zamir, Helen Jiang, Ioannis Brilakis, Martin
  Fischer, and Silvio Savarese.
\newblock 3d semantic parsing of large-scale indoor spaces.
\newblock In {\em Proceedings of the IEEE International Conference on Computer
  Vision and Pattern Recognition}, 2016.

\bibitem{bai2018empirical}
Shaojie Bai, J~Zico Kolter, and Vladlen Koltun.
\newblock An empirical evaluation of generic convolutional and recurrent
  networks for sequence modeling.
\newblock {\em arXiv preprint arXiv:1803.01271}, 2018.

\bibitem{3dr2n2}
Christopher~B Choy, Danfei Xu, JunYoung Gwak, Kevin Chen, and Silvio Savarese.
\newblock 3d-r2n2: A unified approach for single and multi-view 3d object
  reconstruction.
\newblock In {\em Proceedings of the European Conference on Computer Vision
  ({ECCV})}, 2016.

\bibitem{scannet}
Angela Dai, Angel~X. Chang, Manolis Savva, Maciej Halber, Thomas Funkhouser,
  and Matthias Nie{\ss}ner.
\newblock Scannet: Richly-annotated 3d reconstructions of indoor scenes.
\newblock In {\em Proc. Computer Vision and Pattern Recognition (CVPR), IEEE},
  2017.

\bibitem{3dmv}
Angela Dai and Matthias Nie{\ss}ner.
\newblock 3dmv: Joint 3d-multi-view prediction for 3d semantic scene
  segmentation.
\newblock In {\em Proceedings of the European Conference on Computer Vision
  ({ECCV})}, 2018.

\bibitem{gadde2017efficient}
Raghudeep Gadde, Varun Jampani, Renaud Marlet, and Peter Gehler.
\newblock Efficient 2d and 3d facade segmentation using auto-context.
\newblock {\em IEEE transactions on pattern analysis and machine intelligence},
  2017.

\bibitem{sparsecnn}
Benjamin Graham.
\newblock Spatially-sparse convolutional neural networks.
\newblock {\em arXiv preprint arXiv:1409.6070}, 2014.

\bibitem{sparse3dcnn}
Ben Graham.
\newblock Sparse 3d convolutional neural networks.
\newblock {\em British Machine Vision Conference}, 2015.

\bibitem{sparse3dsegmentation}
Benjamin Graham, Martin Engelcke, and Laurens van~der Maaten.
\newblock 3{D} semantic segmentation with submanifold sparse convolutional
  networks.
\newblock {\em CVPR}, 2018.

\bibitem{montecarlo}
P. Hermosilla, T. Ritschel, P-P Vazquez, A. Vinacua, and T. Ropinski.
\newblock Monte carlo convolution for learning on non-uniformly sampled point
  clouds.
\newblock {\em ACM Transactions on Graphics (Proceedings of SIGGRAPH Asia
  2018)}, 2018.

\bibitem{pgm}
Daphne Koller and Nir Friedman.
\newblock {\em Probabilistic Graphical Models: Principles and Techniques -
  Adaptive Computation and Machine Learning}.
\newblock The MIT Press, 2009.

\bibitem{densecrf}
Philipp Kr\"{a}henb\"{u}hl and Vladlen Koltun.
\newblock Efficient inference in fully connected crfs with gaussian edge
  potentials.
\newblock In {\em Advances in Neural Information Processing Systems 24}, 2011.

\bibitem{superpoint}
Loic Landrieu and Martin Simonovsky.
\newblock Large-scale point cloud semantic segmentation with superpoint graphs.
\newblock {\em arXiv preprint arXiv:1711.09869}, 2017.

\bibitem{pointcnn}
Yangyan Li, Rui Bu, Mingchao Sun, and Baoquan Chen.
\newblock Pointcnn.
\newblock {\em arXiv preprint arXiv:1801.07791}, 2018.

\bibitem{4dmrf}
Maria Lorenzo-Vald{\'e}s, Gerardo~I Sanchez-Ortiz, Andrew~G Elkington, Raad~H
  Mohiaddin, and Daniel Rueckert.
\newblock Segmentation of 4d cardiac mr images using a probabilistic atlas and
  the em algorithm.
\newblock {\em Medical Image Analysis}, 8(3):255--265, 2004.

\bibitem{martinovic20153d}
Andelo Martinovic, Jan Knopp, Hayko Riemenschneider, and Luc Van~Gool.
\newblock 3d all the way: Semantic segmentation of urban scenes from start to
  end in 3d.
\newblock In {\em Proceedings of the IEEE Conference on Computer Vision and
  Pattern Recognition}, 2015.

\bibitem{4dseg95}
Tim McInerney and Demetri Terzopoulos.
\newblock A dynamic finite element surface model for segmentation and tracking
  in multidimensional medical images with application to cardiac 4d image
  analysis.
\newblock {\em Computerized Medical Imaging and Graphics}, 19(1):69--83, 1995.

\bibitem{thrust}
Nvidia.
\newblock Thrust: Parallel algorithm library.

\bibitem{surface}
Hao Pan, Shilin Liu, Yang Liu, and Xin Tong.
\newblock Convolutional neural networks on 3d surfaces using parallel frames.
\newblock {\em arXiv preprint arXiv:1808.04952}, 2018.

\bibitem{pytorch}
Adam Paszke, Sam Gross, Soumith Chintala, Gregory Chanan, Edward Yang, Zachary
  DeVito, Zeming Lin, Alban Desmaison, Luca Antiga, and Adam Lerer.
\newblock Automatic differentiation in pytorch.
\newblock 2017.

\bibitem{pointnet}
Charles~Ruizhongtai Qi, Hao Su, Kaichun Mo, and Leonidas~J. Guibas.
\newblock Pointnet: Deep learning on point sets for 3d classification and
  segmentation.
\newblock {\em arXiv preprint arXiv:1612.00593}, 2016.

\bibitem{pointnetpp}
Charles~Ruizhongtai Qi, Li Yi, Hao Su, and Leonidas~J Guibas.
\newblock Pointnet++: Deep hierarchical feature learning on point sets in a
  metric space.
\newblock In {\em Advances in Neural Information Processing Systems}, 2017.

\bibitem{octnet}
Gernot Riegler, Ali~Osman Ulusoy, and Andreas Geiger.
\newblock Octnet: Learning deep 3d representations at high resolutions.
\newblock In {\em Proceedings of the IEEE Conference on Computer Vision and
  Pattern Recognition}, 2017.

\bibitem{ruemonge2014}
Hayko Riemenschneider, Andr{\'a}s B{\'o}dis-Szomor{\'u}, Julien Weissenberg,
  and Luc Van~Gool.
\newblock Learning where to classify in multi-view semantic segmentation.
\newblock In {\em European Conference on Computer Vision}. Springer, 2014.

\bibitem{riemenschneider2014learning}
Hayko Riemenschneider, Andr{\'a}s B{\'o}dis-Szomor{\'u}, Julien Weissenberg,
  and Luc Van~Gool.
\newblock Learning where to classify in multi-view semantic segmentation.
\newblock In {\em European Conference on Computer Vision}, 2014.

\bibitem{synthia}
German Ros, Laura Sellart, Joanna Materzynska, David Vazquez, and Antonio~M.
  Lopez.
\newblock The synthia dataset: A large collection of synthetic images for
  semantic segmentation of urban scenes.
\newblock In {\em The IEEE Conference on Computer Vision and Pattern
  Recognition (CVPR)}, 2016.

\bibitem{splatnet}
Hang Su, Varun Jampani, Deqing Sun, Subhransu Maji, Vangelis Kalogerakis,
  Ming-Hsuan Yang, and Jan Kautz.
\newblock Splatnet: Sparse lattice networks for point cloud processing.
\newblock {\em arXiv preprint arXiv:1802.08275}, 2018.

\bibitem{tangent}
Maxim Tatarchenko*, Jaesik Park*, Vladlen Koltun, and Qian-Yi Zhou.
\newblock Tangent convolutions for dense prediction in {3D}.
\newblock {\em CVPR}, 2018.

\bibitem{segcloud}
Lyne~P Tchapmi, Christopher~B Choy, Iro Armeni, JunYoung Gwak, and Silvio
  Savarese.
\newblock Segcloud: Semantic segmentation of 3d point clouds.
\newblock {\em International Conference on 3D Vision (3DV)}, 2017.

\bibitem{sparsetensor}
Parker~Allen Tew.
\newblock {\em An investigation of sparse tensor formats for tensor libraries}.
\newblock PhD thesis, Massachusetts Institute of Technology, 2016.

\bibitem{3drnn}
Xiaoqing Ye, Jiamao Li, Hexiao Huang, Liang Du, and Xiaolin Zhang.
\newblock 3d recurrent neural networks with context fusion for point cloud
  semantic segmentation.
\newblock In {\em The European Conference on Computer Vision (ECCV)}, September
  2018.

\bibitem{3dmatch}
A. Zeng, S. Song, M. Nießner, M. Fisher, J. Xiao, and T. Funkhouser.
\newblock 3dmatch: Learning the matching of local 3d geometry in range scans.
\newblock In {\em CVPR}, 2017.

\bibitem{stcnn}
Yu Zhao, Xiang Li, Wei Zhang, Shijie Zhao, Milad Makkie, Mo Zhang, Quanzheng
  Li, and Tianming Liu.
\newblock Modeling 4d fmri data via spatio-temporal convolutional neural
  networks (st-cnn).
\newblock {\em arXiv preprint arXiv:1805.12564}, 2018.

\bibitem{crfasrnn}
Shuai Zheng, Sadeep Jayasumana, Bernardino Romera-Paredes, Vibhav Vineet,
  Zhizhong Su, Dalong Du, Chang Huang, and Philip H.~S. Torr.
\newblock Conditional random fields as recurrent neural networks.
\newblock In {\em Proceedings of the 2015 IEEE International Conference on
  Computer Vision (ICCV)}, 2015.

\end{thebibliography}
    }
\endgroup
}

\clearpage
\includepdf[pages=1, pagecommand={}]{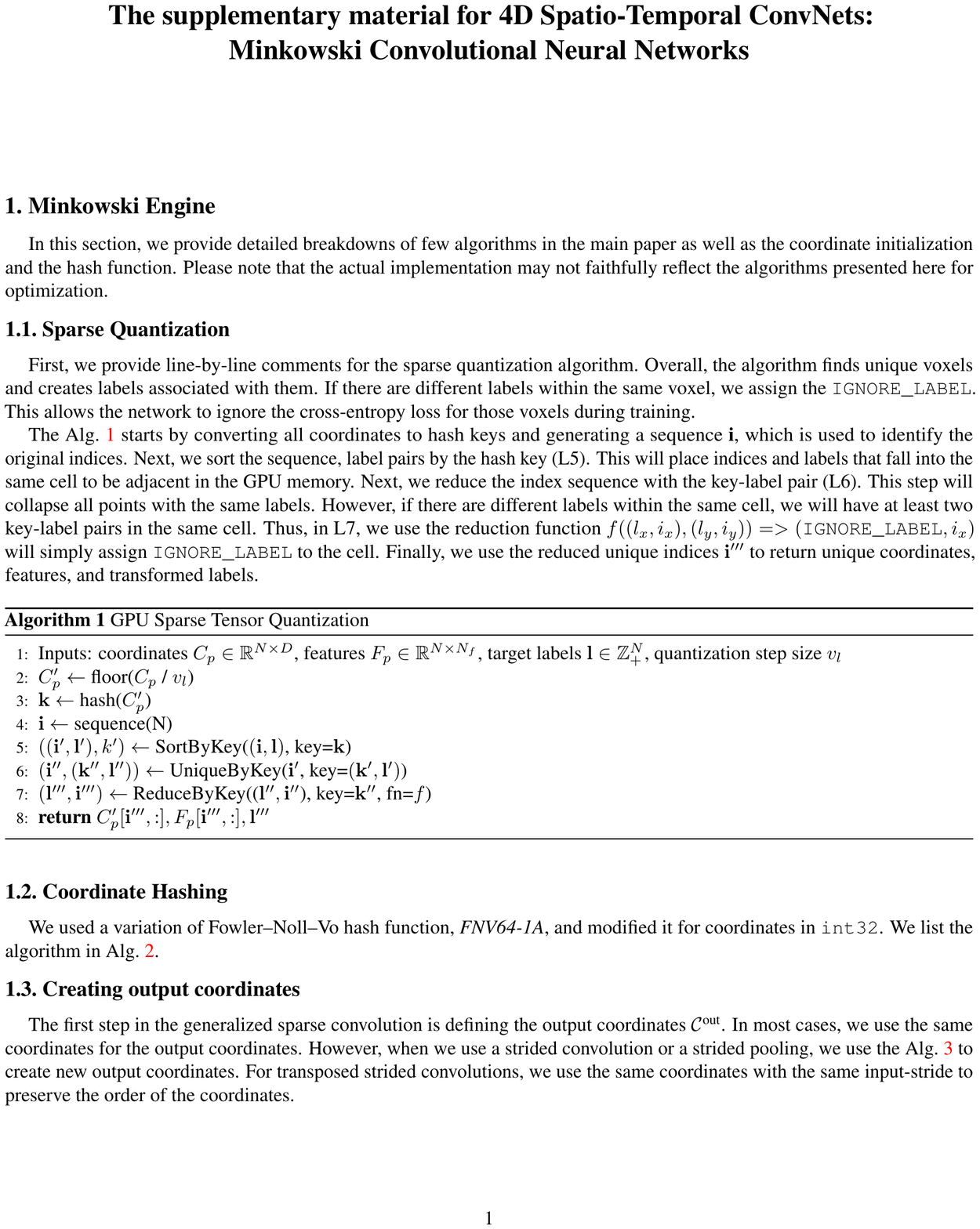}
\includepdf[pages=2, pagecommand={}]{supplementary.pdf}
\includepdf[pages=3, pagecommand={}]{supplementary.pdf}
\includepdf[pages=4, pagecommand={}]{supplementary.pdf}
\includepdf[pages=5, pagecommand={}]{supplementary.pdf}
\includepdf[pages=6, pagecommand={}]{supplementary.pdf}
\includepdf[pages=7, pagecommand={}]{supplementary.pdf}
\includepdf[pages=8, pagecommand={}]{supplementary.pdf}
\includepdf[pages=9, pagecommand={}]{supplementary.pdf}
\includepdf[pages=10, pagecommand={}]{supplementary.pdf}
\includepdf[pages=11, pagecommand={}]{supplementary.pdf}

\end{document}